\documentclass[journal]{IEEEtran}

\usepackage{url,hyperref,lineno,microtype}
\usepackage[onehalfspacing]{setspace}
\usepackage{amsmath,amsthm,amssymb}
\usepackage{graphicx}
\usepackage{import}
\usepackage{color}
\usepackage{tabu}
\usepackage{enumerate}
\usepackage{layouts}
\usepackage{svg}
\usepackage{float}
\usepackage{amsfonts}
\usepackage{booktabs}
\usepackage{siunitx}
\usepackage{wrapfig}
\usepackage{float}

\begin{document}
\onecolumn
\title{Direct Feedback Alignment with Sparse Connections for Local Learning}

\author{\IEEEauthorblockN{Brian Crafton, Abhinav Parihar, Evan Gebhardt, Arijit Raychowdhury}

\IEEEauthorblockA{School of Electrical and Computer Engineering,
Georgia Institute of Technology, Atlanta, GA 30332 USA}}

\maketitle

%%%%%%%%%%%%%%%%%%%%%%%%%%%%%%%%%%%%%%%%%%%%%%%%%%%%%%%%%%%%%%

\begin{abstract}
\noindent
Recent advances in deep neural networks (DNNs) owe their success to training algorithms that use backpropagation and gradient-descent. 
Backpropagation, while highly effective on von Neumann architectures, becomes inefficient when scaling to large networks. 
Commonly referred to as the weight transport problem, each neuron's dependence on the weights and errors located deeper in the network require exhaustive data movement which presents a key problem in enhancing the performance and energy-efficiency of machine-learning hardware.
In this work, we propose a bio-plausible alternative to backpropagation drawing from advances in feedback alignment algorithms in which the error computation at a single synapse reduces to the product of three scalar values.
Using a sparse feedback matrix, we show that a neuron needs only a fraction of the information previously used by the feedback alignment algorithms.
Consequently, memory and compute can be partitioned and distributed whichever way produces the most efficient forward pass so long as a single error can be delivered to each neuron. 
We evaluate our algorithm using standard datasets, including ImageNet, to address the concern of scaling to challenging problems. 
Our results show orders of magnitude improvement in data movement and $2\times$ improvement in multiply-and-accumulate operations over backpropagation.
Like previous work, we observe that any variant of feedback alignment suffers significant losses in classification accuracy on deep convolutional neural networks.
By transferring trained convolutional layers and training the fully connected layers using direct feedback alignment, we demonstrate that direct feedback alignment can obtain results competitive with backpropagation. 
Furthermore, we observe that using an extremely sparse feedback matrix, rather than a dense one, results in a small accuracy drop while yielding hardware advantages.
All the code and results are available under https://github.com/bcrafton/ssdfa. 

\end{abstract}

%%%%%%%%%%%%%%%%%%%%%%%%%%%%%%%%%%%%%%%%%%%%%%%%%%%%%%%%%%%%%%

\section{Introduction} \label {intro}

The demise of Dennard scaling~\cite{dennard1974design} and decline of Moore’s Law~\cite{present2000cramming} have exposed the fundamental scaling limitations of the von Neumann models of computing. This transition is accompanied by the realization that in a fast evolving, socially interconnected world, we are observing a seismic shift in the amount of unstructured data that need to be processed in real-time~\cite{najafabadi2015deep} which has heralded the third wave of Artificial Intelligence and the exponential growth of Machine Learning in data-analytics, real-time control, computer vision, robotics and so on. We expect that intelligent systems of the future will be limited by the energy growth of data movement rather than compute. Therefore, we need fundamentally new approaches to sustain the exponential growth in performance beyond the end of the current road-map. In particular, we observe that new computing models that deal with “data analytics” have compute and storage interleaved in a fine grained manner - not separated as in the von Neumann world. Moving forward, computing technology will heavily penalize separation of data and compute and we need to marry them in better ways to handle emergent applications.

\begin{figure}[t]
\includegraphics[width=\textwidth]{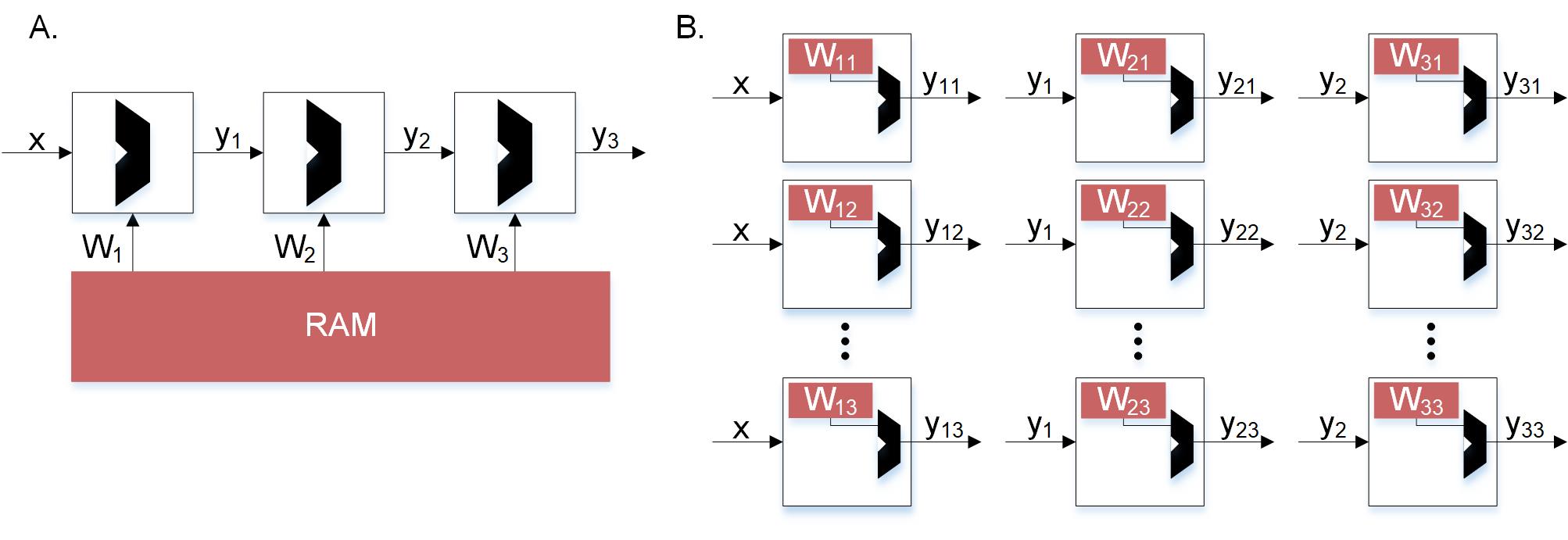}
\caption{Hardware implementations of inference. Comparing von Neumann architecture with distributed memory architecture to avoid bottleneck. \textbf{A}. Inference constrained by von Neumann architectures. In traditional backpropagation, weight updates not only depend on error but also the other weights. This prevents a distributed architecture. \textbf{B.} Unconstrained inference. Using direct feedback alignment, inference can be distributed and parallel because weight updates depend only on the error and random feedback values.}
\label{fig:inference}
\end{figure}

The idea of computing locally on data finds its inspiration from the human brain where local processing and updates is  preferred to global movement of data. Hence, neuromorphic computing seeks to fundamentally improve the power efficiency of cognitive systems by bringing ideas inspired from biology to electronic hardware, while maintaining the high accuracy and performance that statistical methods have  provided. In particular, hardware implementations of deep neural networks (DNNs) and their many variants, either in complementary metal oxide semiconductor (CMOS)~\cite{chen2016eyeriss} or emerging technologies~\cite{li2018efficient}, rely on arrays of spatial processors where near-memory \cite{merolla2014million, chen2016eyeriss, bankman2019always} or in-memory~\cite{chi2016prime} logic computes inference from layers of neurons connected via dense or sparse synaptic connections. This is schematically shown in Figure \ref{fig:inference}.  As opposed to a von Neumann architecture (Figure \ref{fig:inference}.A) where all the synaptic weights for a particular layer must be loaded in the memory to compute the activations of the successive layers, in a distributed implementation (Figure \ref{fig:inference}.B) the weights and logic reside locally and avoid the memory bottleneck. The data movement is minimized, with each neuron computing its activation and sending that information to the next layer. In spite of the success of such spatial processing in the inference mode, such an architecture fails to deliver high efficiency when functions that require global information or weights of multiple layers are implemented. This is particularly evident during the training of DNNs where back-propagation (BP) and stochastic gradient-descent (SGD) have found wide-spread adoption~\cite{lecun2015deep}. In BP with SGD, the transpose of the weights of the deeper layers in a network are needed to compute error gradients and the weight updates of the shallower layers, thus requiring global movement of data. This problem is commonly referred to as the weight transport problem \cite{grossberg1987competitive, lillicrap2016random}. As the networks become deeper to keep up with the complexity of the applications, the weight transport problem becomes exacerbated.

While neural networks require many expensive multiply-and-accumulate (MAC) operations, the cost of data movement is higher \cite{chen2016eyeriss}.
To make the problem worse, for every MAC operation multiple reads and potentially a write are required.
Furthermore, the cost of loading data from off chip DRAM is orders of magnitude more expensive than loading from spatially local on chip memories. 
\cite{kwon2018maestro} show through simulation the breakdown of energy consumption in the different layers of a CNN. 
The majority of the energy consumption comes from memory reads and writes, rather than MAC operations.
The strategy these accelerators employ is data reuse. Because the cost of data movement is so high, it is important that each word of data is reused as many times as possible before being flushed from the cache. 
Optimizing for data reuse, these accelerators can achieve several times better efficiency over data flows that do not use local reuse.
Neuromorphic engineering is another research vector which attempts to minimize data movement by borrowing learning rules from the human brain. For example, in spiking neural networks (SNNs), spike timing dependent plasticity (STDP) has gained popularity because of its local update rule. STDP uses local information, available at a synapse, and has been shown to perform well in unsupervised learning~\cite{diehl2015unsupervised} and supervised learning using Feedback Alignment \cite{neftci2017event, neftci2018data}. 
\cite{davies2018loihi} present a new SNN implementation with tools to perform supervised learning. 
Rather than seeking to optimize current neural network architectures using data reuse, ~\cite{davies2018loihi} uses biological constraints on data movement and calls for new approaches to learning. 
The fundamental constraint is that a weight can only be accessed and modified by its corresponding destination neuron. 
This is further illustrated in Figure \ref{fig:inference}, where each neuron is its own module containing compute and memory. To promote local learning, the weights local to the neuron should not be sent to or from the neuron, only activations and error signals. 
This constraint promotes local learning since the only data movement that occurs are activation signals between adjacent neurons and error computed by the system. 
With this constraint in place, we define data movement as information a neuron must send or receive for each weight update.

One such promising recent work is Feedback Alignment (FA)~\cite{lillicrap2016random}, which has shown that we might be able to bypass the weight transport problem while achieving the same accuracy that BP achieves.
FA uses fixed random feedback weights to propagate the errors back through the layers of a DNN rather than using the actual current network weights to compute the partial error. Consequently, the weights in the shallow layers of the network no longer need information about the weights of all the deeper layers. Building on top of this, \cite{nokland2016direct} proposes Direct Feedback Alignment (DFA), where it was shown that the feedback to shallow layers need not be propagated through all the layers. Instead  the error signal can be fed back to the shallower layers through completely random linear transformations. This further reduces the amount of information required to update the weights in the network. To further describe the weight transport problem and its relationship with local learning, ~\cite{baldi2018learning} describes the concept of a learning channel. The learning channel is a physical way in which information about targets and deep weights are transported in the network. 
Backpropagation uses the forward channel in the backward direction. Using the targets and deep weights we compute partial errors to update the rest of the network. Feedback Alignment instead uses a separate channel to transmit the weights and in doing so avoids the weight transport problem.

Other promising algorithms include target propagation~\cite{lee2015difference} and local error learning~\cite{mostafa2018deep}.
Target propagation bypasses the weight transport problem while still solving the credit assignment problem like the Feedback Alignment algorithms. Instead of computing a loss gradient, a target value is assigned to each feed forward layer using auto encoders. 
Local error learning generates local errors at each layer using linear classifiers with fixed random weights. In doing so, errors at the output of the network do not need to be sent back and instead local objective functions are solved rather than a global one. Local error learning bypasses the weight transport problem, solves the credit assignment problem, and also does not even need to send errors to the hidden layers.
All these algorithms have been shown to perform similarly on benchmarks. While their performance is impressive, they still suffer the same problems when scaled to larger networks. Each of the algorithms fails to match the performance of BP as both the complexity of the network and difficulty of the benchmark increases.
\cite{bartunov2018assessing} highlighted this problem showing that when applied to problems like CIFAR100 and ImageNet, the biological algorithms failed to come close to backpropagation. For a couple benchmarks we do observe considerable degradation due to fully connected layers, however we show this problem comes primarily from convolutional layers. In fact, when trained layers are transferred from backpropagation and the fully connected layers are trained using feedback alignment the performance is similar to that of backpropagation.

Although DFA does not require the error signals to be transformed by the weights of the deeper layers of the network, each neuron requires feedback weights for each error in the network. 
This requires a significant number of computations, memory, and data movement that compromises the locality of the algorithm.
While shallow weights are no longer dependent on deep weights, the amount of data movement and memory required to compute the error at each neuron prohibits a near memory architecture.
In this paper, we show a modified version of DFA, sparse direct feedback alignment (SDFA), where we propose that sparse feedback of the error signals can result in small drop in the network's performance but significantly reduces the computational complexity during learning. 
In an extreme version of SDFA, we demonstrate that even a single error feedback signal can enable the network to learn with a small performance loss.
We call this \textit{single connection SDFA} (SSDFA). We systematically study, through empirical demonstrations, the role of sparsity and rank of the feedback matrix on the network's performance. Our work demonstrates that SSDFA inherits the computational advantages of local learning similar to bio-mimetic networks, while maintaining the high accuracy of BP with SGD (hitherto simply referred to as BP).

\section{Sparse Direct Feedback Alignment} \label {SDFA}

\noindent Direct Feedback Alignment was a remarkable step forward in training fully connected networks. By replacing the backward propagation from deeper layers with a single random matrix, we can avoid the weight transport problem and enable new hardware for training neural networks. 
In small networks, DFA seems feasible since each neuron requires connections to only a few error signals. 
However as the size of the network increases, the size of the feedback matrix also increases and in effect, each weight update needs more information. As an illustrative example, consider a three layered network with 100 hidden neurons which can be trained for classifying the MNIST handwritten digit dataset.
Because each image size is 28 by 28 pixels and there are 10 classes, our network size will be 784 (28x28) - 100 - 10.
In this example, each of the 100 hidden neurons require connections to each of the 10 error signals at the output. 
While this may seem plausible for MNIST, for the AlexNet \cite{krizhevsky2012imagenet} or VGG16 \cite{simonyan2014very} networks that are used to classify the ImageNet dataset, \cite{deng2009imagenet} the number of connections becomes much larger. 
The ImageNet dataset classifies 1000 different classes of images which are re-sized to 224 by 224 pixels \cite{simonyan2014very}. In the case of DFA, each of the 4096 neurons in the first fully connected layer requires a connection to each of the 1000 errors. This quadratic increase in connections prevents DFA from scaling to larger problems without incurring significant computational penalty compared to a more local learning rule.

To relax this problem, we introduce SDFA where the error signal is fed back to all the hidden layer neurons through a highly sparse feedback matrix. SDFA with a sparse feedback matrix enables each neuron to compute its error using a fewer number of error signals. 
Consequently, as we will demonstrate, a hardware design that implements SDFA based learning requires significantly less data movement in the form of error signals, rendering it more efficient both in terms of throughput and power. 
We empirically demonstrate that sparsity plays a negligible role on the network's accuracy as long as the feedback matrix is full-rank or near-full-rank. This leads to SSDFA, which can enable local learning requiring only a single global error to be transferred per neuron, while incurring a small loss of accuracy.

\begin{wrapfigure}{r}{0.4\textwidth}
\includegraphics[width=0.35\textwidth]{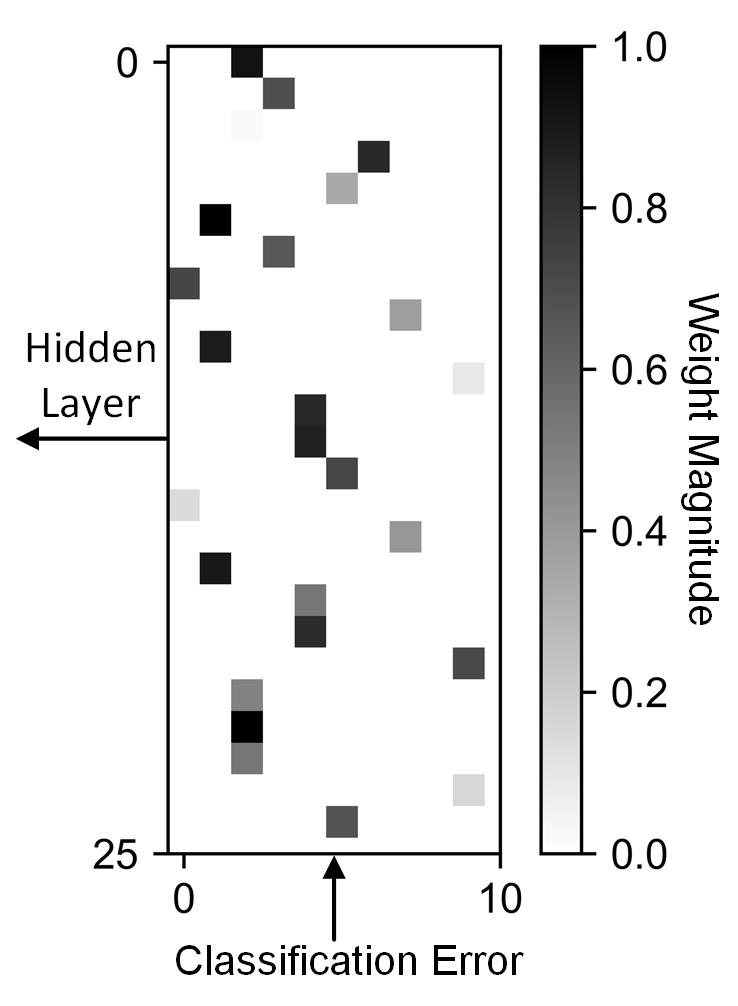} 
\caption{A sparse feedback matrix where each hidden neuron is connected to a single error. Only one of the ten connections between a neuron is non-zero and trainable. }
\label{fig:sparse}
\end{wrapfigure}

We define a feedback as sparse when the values in the feedback matrix that model the connections between the errors and the hidden neurons are mostly zero.
The implication on hardware is that most of the connections do not exist, and therefore, do not require data movement or computation. 
Biological networks share similar properties, where each neuron updates its weights using local values \cite{o2000computational, diehl2015unsupervised}.
In this work, we use the percentage of zero valued connections to quantify sparsity.
In Figure~\ref{fig:sparse} we present an example of a sparse feedback matrix where each of the 25 hidden neurons is connected to one of 10 errors. In this case the sparsity is 90\% because only one of 10 errors are used. We also demonstrate that even an extremely sparse feedback matrix, with 99.9\% sparsity, can be used to achieve high accuracy on the ImageNet dataset. 

\begin{figure}[t]
\includegraphics[width=\textwidth]{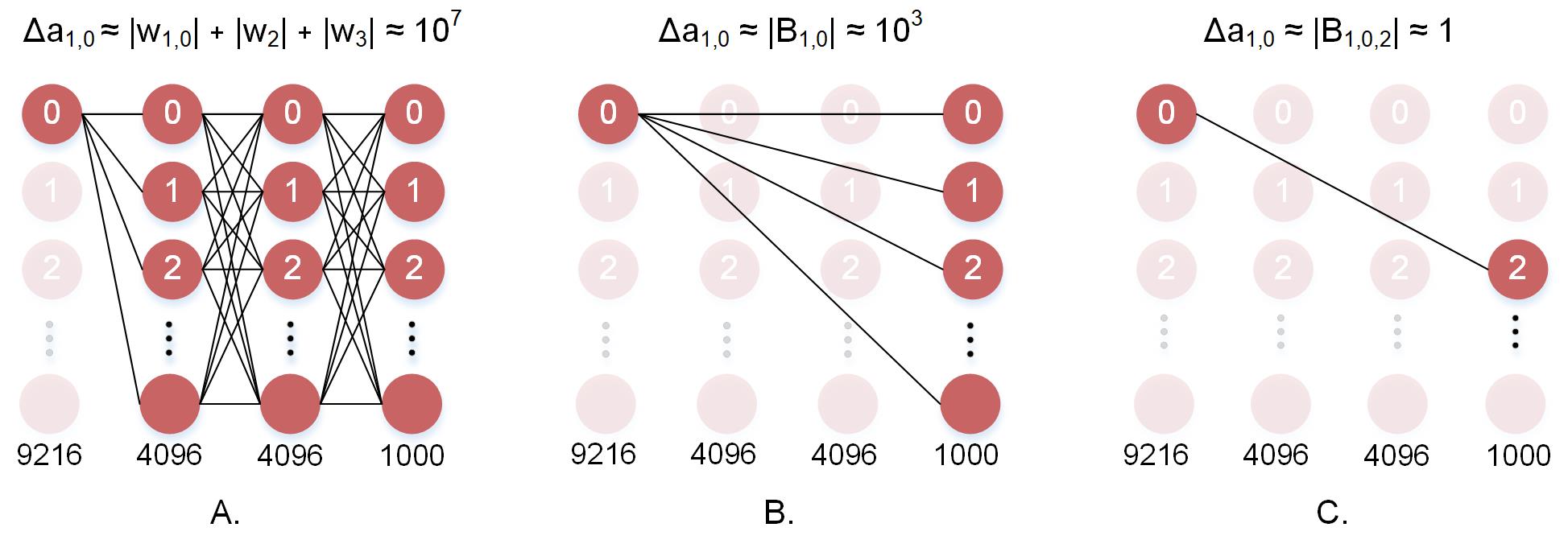} 
\caption{Neuron-level memory dependence of the different algorithms. \textbf{A. Backpropagation:} The error at the first layer is computed using all the weights in the deeper layers. This is the weight transport problem. \textbf{B. DFA:} the error at the first layer is only $e \cdot B$. This solves weight transport, however it still requires 1000 FB weights. \textbf{C. SSDFA:} The error at the first layer is dependent only on a single error and a single feedback weight.}
\label{fig:networks}
\end{figure}

In Figure \ref{fig:networks}, we schematically compare BP, DFA, and SDFA for a prototypical fully connected network consisting of four layers of neurons. Figure \ref{fig:networks} shows the number of connections the pre-synaptic neuron in the first hidden layer needs to compute a weight update. In BP, the neuron is dependent on each and every neuron it has a direct or indirect connection to. Hence, the weight update of the synapse is dependent on all model weights deeper in the network. The number of connections grows as the network gets deeper, leading to the weight transport problem. 
On the other hand, in DFA, the neuron is dependent only on the number of errors of the network. While this decouples the forward and backward pass and relieves the weight transport problem, complex networks with many output neurons will still have a large number of feedback connections. 
In the proposed SDFA, a weight update is dependent only on the few errors it is connected to. Consequently, even if the network scales in both depth and complexity, the number of connections for updating a synaptic weight is low. In the proposed extreme scenario for SDFA, we name SSDFA, each neuron receives only one error signal and can successfully update its weights with a single error signal.

\begin{figure}[t]
\includegraphics[width=\textwidth]{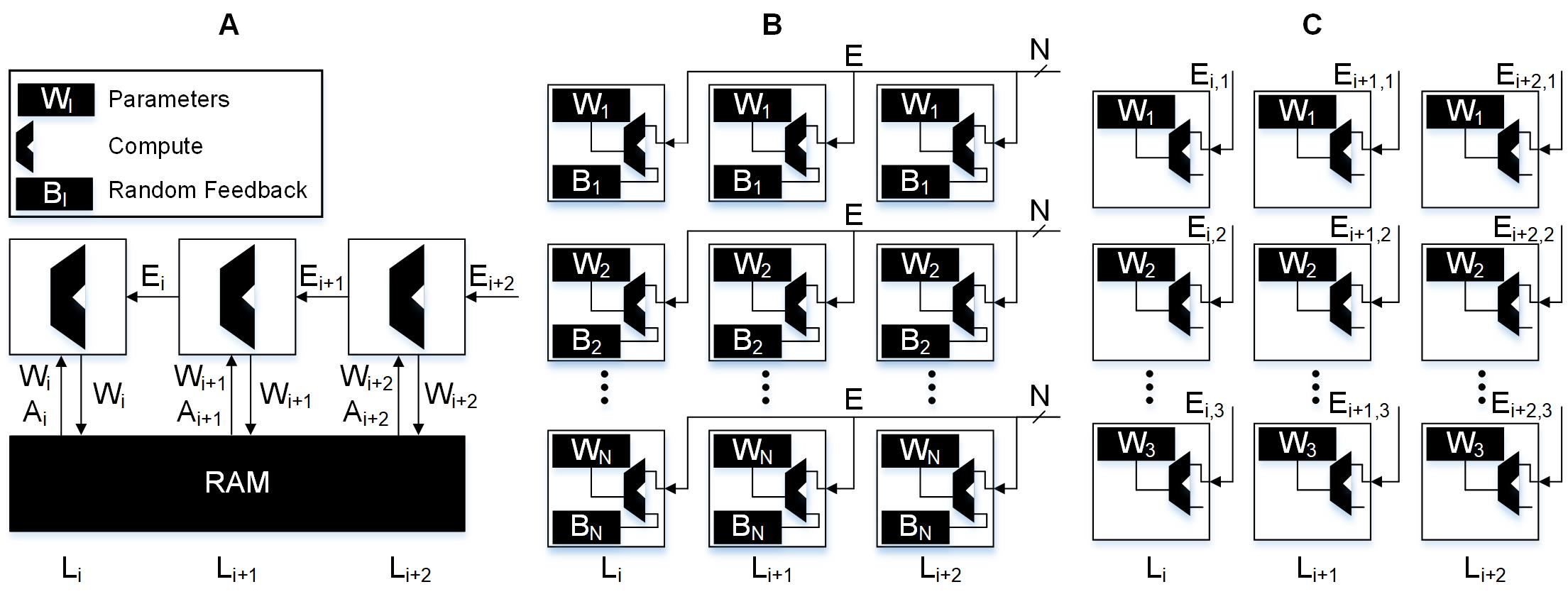} 
\caption{Data movement through the substrate of different algorithms. \textbf{A. Backpropagation:} In traditional von Neumann architectures weights and activations from the forward pass must be accessed from main memory. The majority of data transfer occurs when moving the large weight matrices from main memory to compute. \textbf{B. DFA:} In a local learning implementation only the error vector needs to be sent to the neurons. In this case the neuron must receive all $N$ errors and store an additional $N$ random feedback weights. \textbf{C. SSDFA:} In the single sparse connection implementation of DFA, only a single error needs to be sent to each neuron and only a single random feedback constant needs to be stored. This reduces the bandwidth requirement and feedback weight storage by a factor of $N$.  }
\label{fig:backward}
\end{figure}

Using only a single error per neuron, we are able to greatly reduce the amount of data movement in the backwards pass. 
While there will be significant savings in memory accesses and MAC operations, the key improvement is in data movement. 
However, in order to take advantage of this, a von Neumann architecture is not an ideal choice for bio-mimetic algorithms. In particular, recent advances in spatial array processors with near-memory computing can achieve significant advantages in both performance and energy-efficiency~\cite{chen2017eyeriss, hsu2014ibm}.
In Figure~\ref{fig:backward} we show a comparison of the different algorithms and the data movement required in the backwards pass.
We illustrate in Figures \ref{fig:backward}.A to \ref{fig:backward}.C how the data movement greatly reduces. 
With backpropagation on a von Neumann architecture (Figure \ref{fig:backward}.A) we must read the weights and activations to and from the main memory. DFA on a neuromorphic architecture (Figure \ref{fig:backward}.B) relaxes the weight transport problem and keeps the read, write, and MAC operations local to the neuron itself with the only data movement occurring when the error is sent backwards. Lastly, in Figure \ref{fig:backward}.C the error sent backwards is only a single scalar, reducing data movement in the backwards pass to its minimal form. 
In Table \ref{tab:compare}, we compare the number of reads, writes, MACs, and data movement across the different algorithms as a function of the neurons and weights in the network. $|A|$, $|W|$, $|B|$, and $|E|$ refer to the number of neurons, weights, feedback weights, and errors, respectively. We also use $|b|$ and $|e|$ to refer to the reduced number of weights and errors from the sparse feedback matrix. 

\begin{table}[t]
\begin{tabular}{ p{3cm} p{3cm} p{3cm} p{3cm} p{3cm} }
\midrule
Method & Reads              & Writes      & MACs         & Movement                 \\
\midrule
BP     & $|W| + |A|$        & $|W|$       & $2|W|$       & $|W| + |A| + |E_{i+1}|$  \\
DFA    & $|W| + |B|$        & $|W|$       & $|W| + |B|$  & $|E|$                    \\
SDFA   & $|W| + |b|$        & $|W|$       & $|W| + |b|$  & $|e|$                    \\
\bottomrule
\end{tabular}
\caption{Data Movement Comparison of Error Assignment Algorithms}
\label{tab:compare}
\end{table}

%%%%%%%%%%%%%%%%%%%%%%%%%%%%%%%%%%%%%%%%%%%%%%%%%%%%%%%%%%%%%%

\section{Mathematical Formulation} \label {results}

The primary contribution of this work is to show the benefits of using sparse feedback that can eventually enable local learning in neuromorphic hardware implementations. Empirically, we show on standard networks and datasets that sparsity, even extreme-sparsity (SSDFA), results in negligible loss of accuracy while reducing data movement during training by orders of magnitude.

We investigate BP, DFA, and SDFA for the fully connected network architecture. The feed forward computation can be written as
\begin{gather} 
\label{inf1}
y_1 = W_1 \cdot x,   a_1 = f(y_1) \\
\label{inf2}
y_2 = W_2 \cdot a_1,   a_2 = f(y_2) \\ 
\label{inf3}
y_n = W_n \cdot a_{n-1},   a_n = f(y_n)
\end{gather}
\noindent 
where $x$ is the feature vector and $W_i$ is the weight matrix connecting layer $i-1$ to layer $i$ $(y_0 = x)$. The dot product of $x$ and $W_i$ yields $y_i$, and applying the non-linear activation function $f$ results in the activation at layer $i$, $a_i$. 
Each of these algorithms computes the error at a specific layer. 
The error at the last layer of the network, $n$, is the classification error $e$. 
BP computes the error at each hidden layer \textit{l}, $\delta a_l$, by transposing the weight matrices W and multiplying by the gradient of the activation function. 
These layerwise computations for BP can be written as 
\begin{gather} 
\label{bp1}
\delta a_n = e \\
\label{bp2}
\delta a_2 = W_2^T \cdot \delta a_{3} \odot f^{'}(a_2) \\
\label{bp3}
\delta a_1 = W_1^T \cdot \delta a_{2} \odot f^{'}(a_1) 
\end{gather}
\noindent 
where $ \odot $ is the element-wise multiplication operator. BP requires all the deeper weights in the network in order to compute the error at a layer earlier in the network.
DFA bypasses this dependence, only requiring a random matrix B to be multiplied with the error vector.
The layerwise error computations for DFA are
\begin{gather} 
\label{dfa1}
\delta a_n = e \\
\label{dfa2}
\delta a_2 = B_2 \cdot e \odot f^{'}(a_2) \\
\label{dfa3}
\delta a_1 = B_1 \cdot e \odot f^{'}(a_1) 
\end{gather}
where we observe that the error at each layer does not depend on the error at any other layer.
In the proposed SDFA, the error computation is identical to Equations (\ref{dfa1}), (\ref{dfa2}), and (\ref{dfa3}), except with an added constraint that B is sparse. At the neuron level this represents an important detail in the physical implementation of neuromorphic hardware. In DFA the error computation for a neuron $i$ in layer $l$ is
\begin{equation} 
\label{dfa_neuron}
% \Delta A_{i} = \sum_{j=1}^N{e_j}{B_{ij}} \odot f^{'}(A_{i})
\delta a_{l, i} = B_{l, i} \cdot e \cdot f^{'}(a_{l, i})
\end{equation}
which is the inner product of two vectors multiplied by a scalar.
This is computationally challenging for complex networks. For example, when we use VGG16 network on the ImageNet dataset, the number of MAC operations for each neuron is 1000 corresponding to the number of output classes. On the other hand, in SDFA, when $B$ is sparse, the number of MACs required for each update will be significantly less, depending on the number of non-zero entries in the corresponding row of $B$. In an extreme scenario, when each row of $B$ has only 1 non-zero entry, the error computation reduces to
\begin{equation} 
\label{sparse_neuron}
\delta a_{l,i} =  B_{l,i,j} \cdot e_j \cdot f^{'}(a_{l,i})
\end{equation}
which is the product of three scalars. 

The significance of this comes from past work in local learning and bio-plausible 3-factor learning rules. 
Locality is a constraint on the learning rule, and for a learning rule to be local each of the variables used in the learning rule must also be local. 
\cite{baldi2016theory} simplifies learning rules to the following forms
\begin{gather}
\label{baldi_a}
\delta W_{i,j} = f(o_i, o_j, W_{i,j}) \\
\label{baldi_b}
\delta W_{i,j} = f(o_i, o_j, W_{i,j}, t_{i,j})
\end{gather}
where $o_i$ and $o_j$ are the output values of neurons $i$ and $j$ connected by weight $W_{i,j}$. The variable $t_{i,j}$ is the target value computed for weight $W_{i,j}$. 
An example of Equation (\ref{baldi_a}) is Hebbian learning \cite{hebb2005organization} using Oja's rule \cite{oja1982simplified}. 
In Oja's rule the synapse uses only the activations of the neurons that it is connected to, and its own state variable. 
Commonly referred to as the \textit{3 factor learning rule}~\cite{baldi2016theory}, Equation (\ref{baldi_b}) requires a target value.
Each of the algorithms discussed in this work (BP, DFA, SSDFA), can be simplified to Equation (\ref{baldi_b}).
In a supervised learning problem, computing the target value requires information about the network error.  
By the definition we gave earlier, we claim that data movement is information a neuron must send or receive. 
Therefore, any information sent or received by the neuron is non-local, which implies that error information in the backwards pass is non-local. 
Furthermore, we realize that any supervised learning algorithm would be non-local since all error information is not local to the neuron.
As a result, none of the learning rules discussed in this paper exhibit pure locality, which would only be possible using a variant of Oja's rule given this definition. 
Although some recent work \cite{mostafa2018deep} has shown a method of local supervised learning, for practical problems the learning rule fails to remain local.
Local Error Learning \cite{mostafa2018deep} locally stores the labels at each layer and computes the prediction error using a random matrix at that layer. 
Since the prediction error is computed locally, it no longer needs to be transported from the end of the network to the given layer. 
In theory this prevents data movement, but for practical problems it is not possible to store data labels in this way.
Two examples of systems where this technique fails to scale are real time systems and very large datasets. 
For real time systems like an autonomous drone, we do not have the labels to store locally in each layer. 
These labels would need to be transported from main memory, which is more expensive than moving errors from the last layer of the network.
A similar problem arises for very large datasets. 
For very large datasets like ImageNet, we would need to store over one million labels locally in each layer or neuron. 
Storing this amount of information is not possible to do while maintaining locality. 

\indent The algorithms discussed in this paper have different levels of locality, which depend on the amount of non-local information required to compute the target value. 
To compute the target value for BP, we require all the downstream weights and errors. By computing the errors layer by layer, we are able to avoid redundant computation.
However, this still requires the most data movement. 
In the case of DFA, we can store the random feedback weights local to the neurons themselves so the only data movement required are the downstream errors. As we scale DFA to larger datasets with more errors and therefore larger feedback matrices, we further compromise the locality of the algorithm because we cannot store large feedback matrices and route hundreds of errors to a single neuron. 
When the feedback matrix is made sparse (SDFA), we can scale the size of the network without requiring dense feedback connections. 
When only a single feedback connection is used (SSDFA), the target value is a function of only a single error and a single feedback weight. Although SSDFA uses non-local information, it is minimal in the sense that each neuron needs to receive only a single error. 

%%%%%%%%%%%%%%%%%%%%%%%%%%%%%%%%%%%%%%%%%%%%%%%%%%%%%%%%%%%%%%%%%%%%%%%%%

\section{Results}

To evaluate the performance of SDFA and SSDFA, we benchmark the proposed algorithms on standard vision datasets vis-a-vis BP and DFA. We empirically evaluate the accuracy of the models with respect to the desired characteristics of the feedback matrix.

\subsection{Benchmarks and Network Architectures}

In our simulations we use 4 standard benchmarks of varying complexity that have been used in prior work. These benchmarks are MNIST~\cite{lecun1990handwritten}, CIFAR10, CIFAR100~\cite{krizhevsky2009learning}, and ImageNet~\cite{deng2009imagenet}. ImageNet is by far the most exhaustive dataset, and like \cite{bartunov2018assessing}, we find that using this benchmark reveals scaling issues related to the Feedback Alignment algorithms. 
For each of these benchmarks we use two networks. 
For MNIST, CIFAR10, and CIFAR100 we use one fully connected network and one convolutional network which are further described in Table ~\ref{tab:arch}. For ImageNet we use two standard convolutional networks: AlexNet~\cite{krizhevsky2012imagenet} and VGG16~\cite{simonyan2014very}, which have produced high accuracy with BP. For each dataset, the corresponding model architectures are summarized in Table~\ref{tab:arch}.

\newcommand{\mnistFC}{FC 400 \newline Softmax 10}
\newcommand{\cifarFC}{FC 1000 \newline FC 1000 \newline FC 1000 \newline Softmax (10, 100)}
\newcommand{\mnistConv}{Conv (3x3 1x1 32) \newline Conv (3x3 1x1 64) \newline Pool (2x2 2x2) \newline FC 128 \newline Softmax 10}
\newcommand{\cifarConv}{Conv (5x5 1x1 96) \newline Pool (3x3 2x2) \newline Conv (5x5 1x1 128) \newline Pool (3x3 2x2) \newline Conv (5x5 1x1 256) \newline Pool (3x3 2x2) \newline FC 2048 \newline FC 2048 \newline Softmax (10, 100)}

\begin{table}[h]
\caption{Network Architectures}
\label{tab:arch}
\begin{tabular}{ p{3cm} p{6cm} p{7cm} }
\midrule
Dataset           & Fully Connected Network & Convolutional Network \\
\midrule
MNIST             & \mnistFC{}              & \mnistConv{}                            \\
\midrule
CIFAR (10, 100)   & \cifarFC{}              & \cifarConv{}                            \\
\midrule
ImageNet          & -                       & Alexnet (\cite{krizhevsky2012imagenet}) \\
\midrule
ImageNet          & -                       & VGG16   (\cite{simonyan2014very})       \\
\bottomrule
\end{tabular} \par
\bigskip
The format for Convolutional layers is as follows: Conv (kernel size, number output channels, stride size). The format for Pool layers is as: Pool (kernel size, stride size). The networks used for the CIFAR10 and CIFAR100 datasets were both derived from~\cite{nokland2016direct}.
\end{table}

\subsection{SDFA and SSDFA for Fully Connected Network}

We start by reporting our findings on fully connected networks. The three properties of the feedback matrix that we study are: 
\begin{enumerate}
    \item \textbf{Rank:} The number of linearly independent vectors in the feedback matrix.  
    \item \textbf{Sparsity:} The percentage of zero-valued weights in the feedback matrix connecting each hidden layer to the classification error. We are particularly interested in the case where each hidden neuron is connected to a single error (SSDFA).
    \item \textbf{Angle:} The angle between the vectorized weight matrix ($W$) and the vectorized feedback matrix ($B$). 
\end{enumerate}

The evaluation platform is setup in TensorFlow \cite{abadi2016tensorflow}. For MNIST and CIFAR10 there are 10 output neurons and hence 10 error signals.
For each combination of rank and sparsity we simulate the models 10 times and statistically observe the accuracy of the network.
To ensure that the network has feedback from all ten errors, the product of rank and connectivity (1 - sparsity) must be equal to or greater than 1. 
In order to create a $N \times M$ feedback matrix connecting $N$ hidden neurons and $M$ errors with rank $R$ and sparsity $S$, we need $R$ linearly independent vectors with sparsity $S$.
When the rank of the matrix is not full, some of the vectors must be linearly dependent. 
As a result, they must have the same zero valued indices. 
When the rank and connectivity product is less than 1, some columns in our matrix will be completely zero. 
This implies these errors will never be propagated backwards and the error at the layer will not depend on these classes. 
Consequently, we are unable to generate results for cases where the product is less than 1. If the network does not have feedback from all ten errors, then the network's performance is impacted, not because of lower rank or higher sparsity, but because only some of the error signals are being propagated back. 

\begin{figure*}[t]
\includegraphics[width=\textwidth]{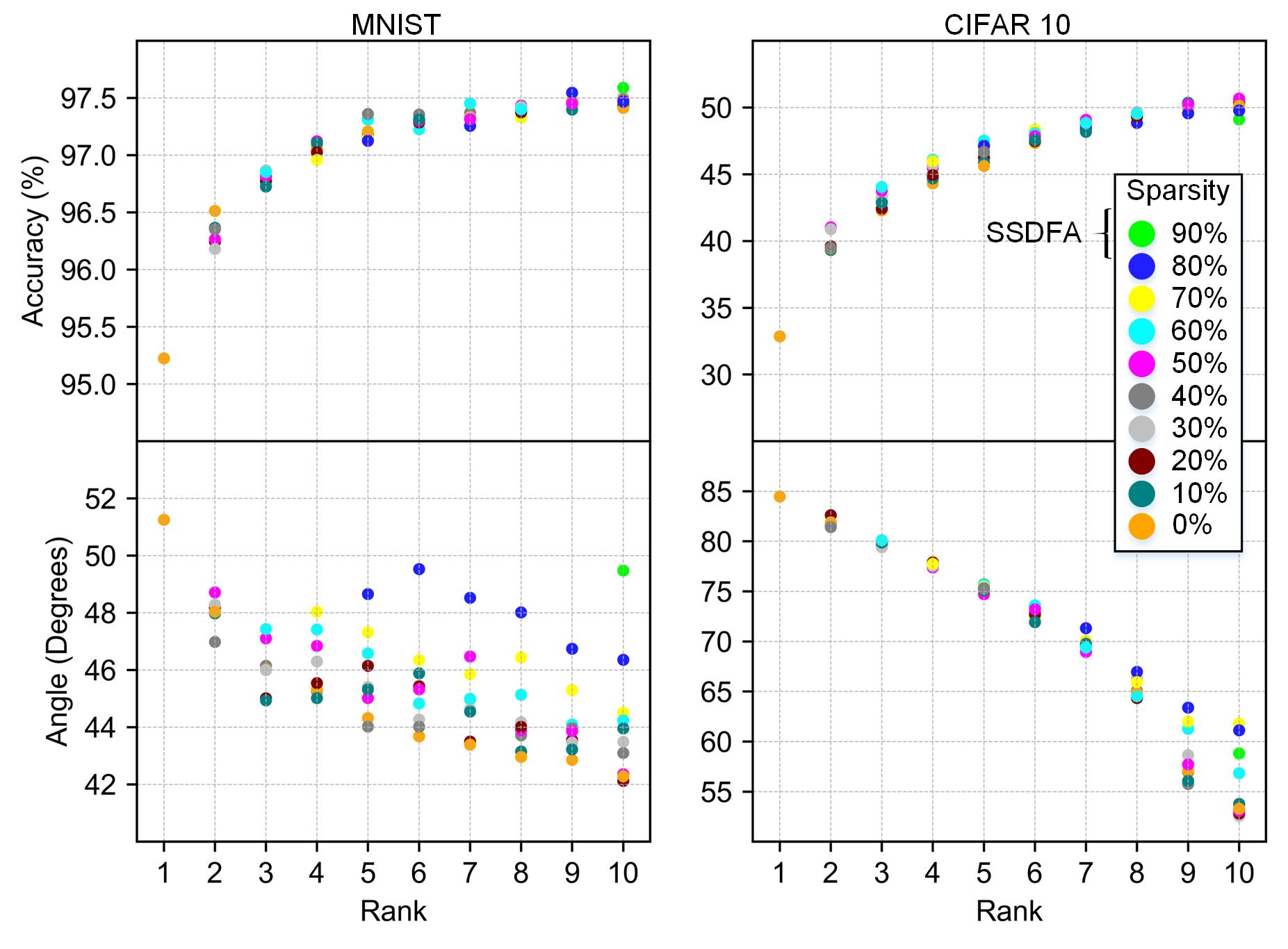}
\caption{Accuracy and angle (in degrees) versus rank for MNIST and CIFAR10 fully connected networks. Data points are grouped by sparsity, and averaged over 10 different simulations.}
\label{fig:plot1}
\end{figure*}

\noindent \textbf{Rank:} Empirically, we observe that the rank of a feedback matrix has the largest impact on the resulting accuracy of the network. As we increase the rank, the accuracy of the network increases  and finally saturates. In Figure~\ref{fig:plot1} we show our results for the MNIST and CIFAR10 datasets. In Figure~\ref{fig:plot1}.A and \ref{fig:plot1}.B, we show the accuracy and angle versus the rank of the feedback matrix, for varying sparsity. For the MNIST dataset, we observe that the test accuracy saturates, as expected at 97.5\%, and is maximum for the full-rank matrix. However for CIFAR10, the accuracy continues to increase as a function of rank without saturation. This shows that the rank of the feedback matrix is a critical design parameter and the feedback needs to be a full-rank matrix to maximize the network's accuracy.

\begin{figure*}[t]
\includegraphics[width=\textwidth]{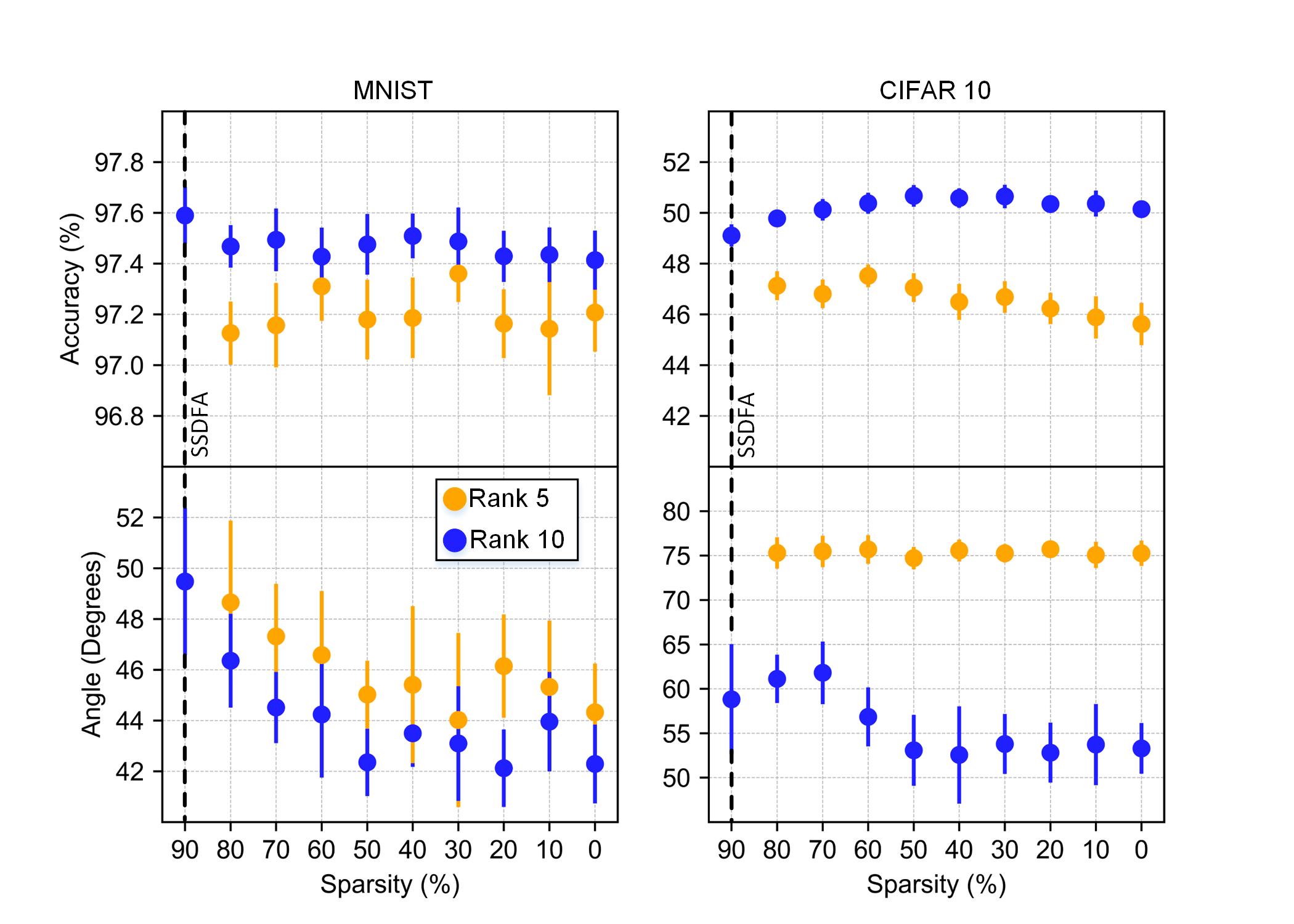}
\caption{Accuracy and angle versus sparsity. Results from rank 5 and rank 10 are shown with bars showing the standard deviation for 10 different simulations. }
\label{fig:plot2}
\end{figure*}

\noindent \textbf{Sparsity:} Our results show that sparsity of the feedback matrix has very little impact on the resulting accuracy of the network in both the MNIST and CIFAR10 networks (Figure \ref{fig:plot1}).
We also observe that sparsity has a negligible impact on the angle between the feedback matrix used and the resulting feed forward weights (Figure \ref{fig:plot2}). 
For many resource constrained systems, a small difference in accuracy for large improvements in power and performance is an excellent trade-off. 
We observe that training with highly sparse feedback matrices, even with just a single feedback error, performs very well -- while significantly reducing the computational demand, as we describe later.

\noindent
\textbf{Angle:} \cite{lillicrap2016random} shows that at each hidden layer, the angle between error gradient computed by FA and BP decreases as the network is trained.
Instead we look at the angle between the vectorized weight matrix and feedback matrix after training.
In the case of propagating errors to shallow layers of the network, the corresponding weight matrix used for the angle calculation is the product of all the weights matrices following this layer. Hence, for the angle calculation for a layer $l$ in a $n$ layer network, the angle is given by

\begin{equation} 
\label{angle}
\angle(B_l, W)  = \angle(B_l, \prod_{i=l+1}^{n} W_{i})
\end{equation}

\noindent
In Figure \ref{fig:plot1} we show a strong correlation of accuracy and angle between the feedback matrix and the weight matrix. This correlation is more apparent for CIFAR10 because it does not plateau as it approaches a full-rank matrix.
This correlation is also illustrated in Figure \ref{fig:plot2} because neither the angle nor the accuracy changes significantly as a function of sparsity. 

In Table~\ref{tab:res_tab1} we summarize the results for fully connected networks. We show the results collected from running BP, DFA, and SSDFA on MNIST, CIFAR10, and CIFAR100 with fully connected networks of different sizes summarized in Table~\ref{tab:arch}.
All the network parameters and code to run experiments can be found under: https://github.com/bcrafton/ssdfa.
The results yielded by BP and DFA are similar to previous work~\cite{nokland2016direct}, which shows that DFA performs similarly to BP for the fully connected networks.
Also like previous work \cite{nokland2016direct}, we observe a non-negligable drop in accuracy for the CIFAR100 benchmark using both DFA and SSDFA. Since both DFA and SSDFA fail to match backpropagation for this benchmark, we infer that the issue is likely caused by the direct feedback and not the sparsity.

\begin{table}[t]
\begin{tabular}{ p{5cm} p{3cm} p{3cm} p{3cm}  }
\midrule
Benchmark & BP   & DFA  & SSDFA      \\
\midrule
MNIST     & 98.2 & 97.8 & 97.5       \\
CIFAR10   & 59.9 & 58.9 & 58.6       \\
CIFAR100  & 33.1 & 29.4 & 28.8       \\
\bottomrule
\end{tabular}
\caption{Test Accuracy (in \%) for Fully Connected Networks}
\label{tab:res_tab1}
\end{table}

\subsection{SDFA and SSDFA for Convolutional Neural Networks}

\noindent Earlier work \cite{nokland2016direct} show that DFA can be used in convolutional neural networks (CNNs). 
However, we note, similar to~\cite{bartunov2018assessing}, as we increase the complexity of our network and dataset, DFA fails to match the accuracy of BP.
For the convolutional network benchmarks, we show our results in Table \ref{tab:res_tab2}. 
Consistent with previous work, the CNNs we use for MNIST show similar performance for BP, DFA, and SSDFA.
However as the problem and network complexities scale to CIFAR10, CIFAR100, and ImageNet, the gap between BP and DFA grows to where their performance is no long comparable.

\begin{table}[t]
\begin{tabular}{ p{5cm} p{3cm} p{3cm} p{3cm} }
\midrule
Benchmark & BP & DFA & SSDFA            \\
\midrule
MNIST              & 99.1 & 98.9 & 98.8 \\
CIFAR10            & 79.6 & 72.3 & 73.1 \\
CIFAR100           & 51.0 & 43.7 & 41.8 \\
ImageNet (Alexnet) & 53.6 & 6.2  & 2.8  \\
\bottomrule
\end{tabular}
\caption{Test Accuracy (in \%) for the Convolutional Neural Networks. Accuracy results are slightly less for ImageNet than originals because minimal pre-processing and data augmentation techniques were used. All results are reported as Top-1 accuracy.}
\label{tab:res_tab2}
\end{table}

\begin{figure}[t]
\centering
\includegraphics[width=\textwidth]{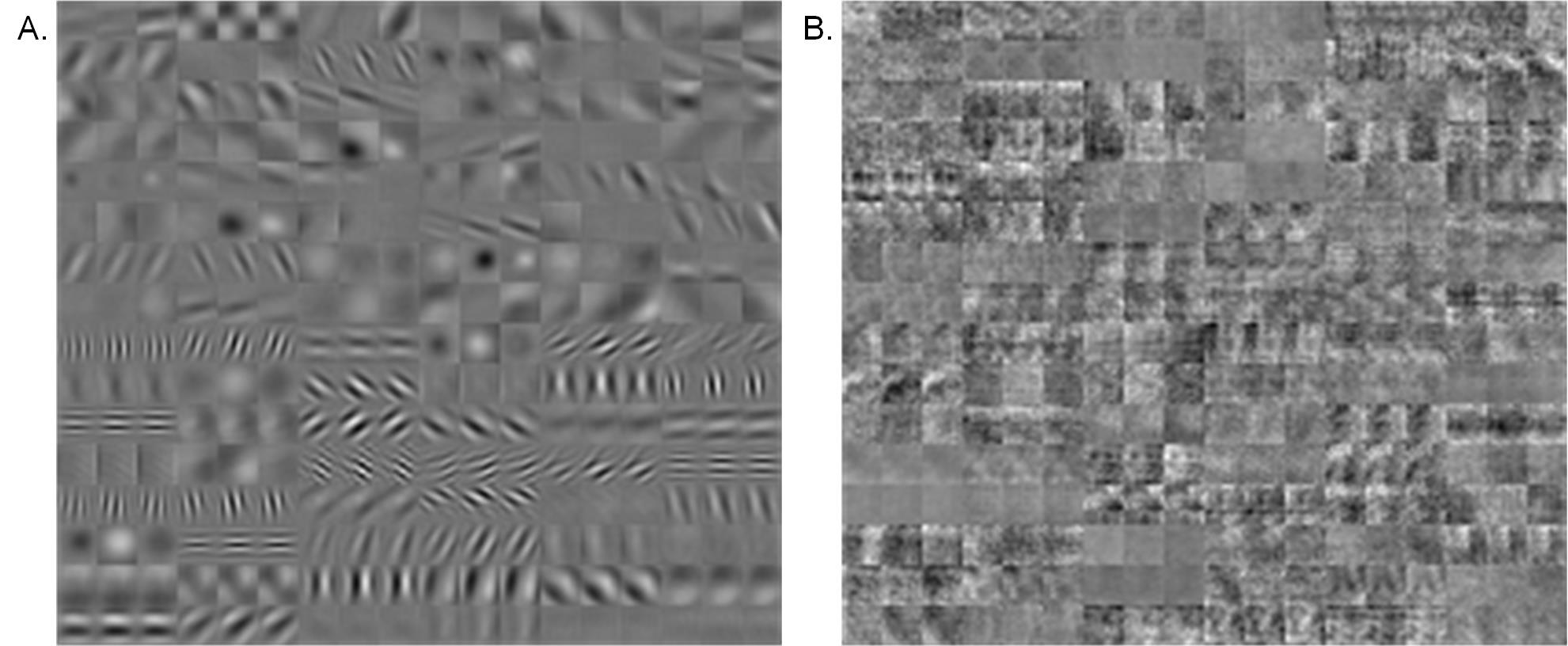}
\caption{Filters acquired from training Alexnet with BP (A) and DFA (B). Filters for BP show shape and spatial structure, while filters from DFA are random.}
\label{fig:filters}
\end{figure}

We attribute this problem to the convolutional layers in the network. 
The convolutional architecture introduces extra constraints on the weight space, and as such, results in stronger constraints on the feedback matrix which can be used.
This can be seen in Figure \ref{fig:filters}, where we show the filter weights for the first convolutional layer from training AlexNet on ImageNet using BP and DFA. 
The resulting filters show completely different patterns. 
The BP filters show a well defined spatial structure, while the filters from DFA are noisy with much less structure. 
From this, we infer that DFA cannot learn convolutional filters as efficiently as BP for large and complex networks.

One way to alleviate this problem is Transfer Learning \cite{pan2010survey}.
Transfer Learning is a technique where a model from a prior task is used to initialize a model for another task. This is of particular importance in mobile platforms where pre-trained models are transferred to the mobile platform and the models are further refined via on-device learning.
By borrowing the weights from the convolutional layers of CNNs which yield good results, we can bypass training these layers with DFA or SSDFA. 
We demonstrate that by reusing these convolutional filters and training the fully connected layers using DFA and SSDFA, we can achieve similar performance to BP. 
In Table \ref{tab:res_tab3}, we show the results for five different benchmarks of varying complexity. 
In each of these benchmarks, we transfer the weights from CNNs trained with BP and train the fully connected layers of each network.
In all of these benchmarks, the performance of DFA and SSDFA is competitive with BP. The largest performance degradation we observe is 2.7\% on the ImageNet benchmark using the AlexNet network.

\begin{table}[t]
\begin{tabular}{ p{5cm} p{3cm} p{3cm} p{3cm} }
\midrule
Benchmark          & BP   & DFA  & SSDFA      \\
\midrule
MNIST              & 99.1 & 99.1 & 99.0       \\
CIFAR10            & 77.1 & 77.8 & 78.0       \\
CIFAR100           & 48.2 & 49.0 & 48.2       \\
ImageNet (Alexnet) & 49.0 & 48.8 & 46.3       \\
ImageNet (VGG)     & 65.8 & 65.3 & 64.5       \\
\bottomrule
\end{tabular}
\caption{Test Accuracy (in \%) for after Transfer Learning of the Convolutional Filters followed by SSDFA for the Fully Connected Layers. Accuracy results are slightly less for ImageNet than originals because minimal pre-processing and data augmentation techniques were used. All results are reported as Top-1 accuracy.}
\label{tab:res_tab3}
\end{table}

\subsection{Computational Advantage}

\noindent
The primary advantage of using SSDFA is that it greatly reduces data movement in the backwards pass. 
Furthermore, SSDFA also reduces the number of multiply-and-accumulate operations (MACs) and memory-reads. 
Computationally, this is motivated by biology where memory and compute are interleaved and global movement of data is minimal. 
New neuromorphic architectures for deep learning~\cite{lee2018unpu, shin201714, lee20197} and reinforcement learning~\cite{cao201914, kim20192, amaravati201855, amravati201855nm} seek to apply this constraint to avoid the communication overhead. 
However, as we have noted earlier, BP violates this constraint. 
Rather than requiring local information, it requires information from the weights of the deeper layers in the network. 
By decoupling the forward and backward weights, DFA and SSDFA do not require information about the weights deeper in the network.

\begin{figure*}[t]
\includegraphics[width=\textwidth]{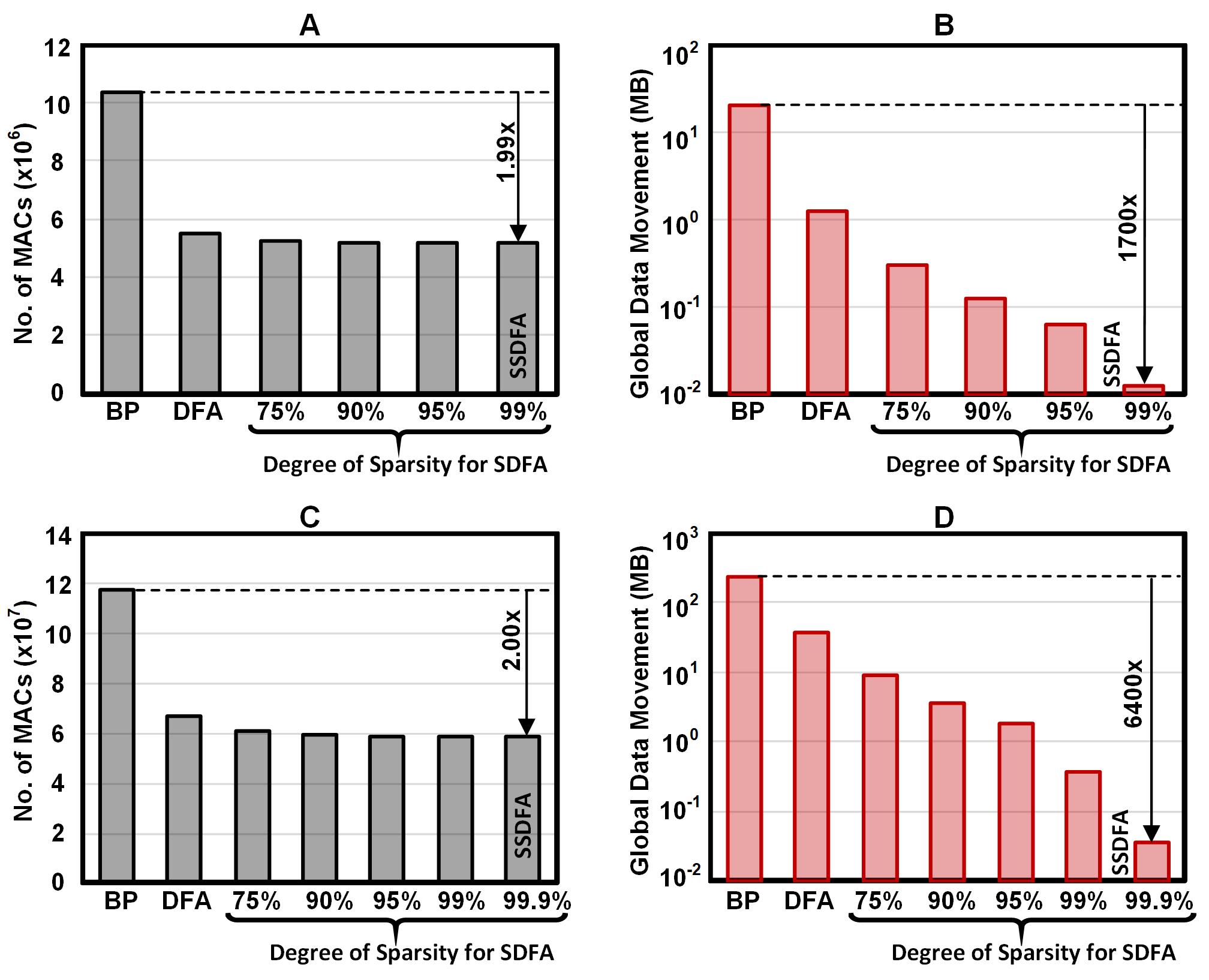}
\caption{ Total number of MACs and Data Movement (in MB) for a single training example on CIFAR100 (\textbf{A, B}) and ImageNet (\textbf{C, D}) across BP, DFA, and SDFA }
\label{fig:ops}
\end{figure*}

\indent In Figure \ref{fig:ops}, we show the MAC and data movement savings when implemented in a near memory architecture on the CIFAR100 and ImageNet datasets.
In Section \ref{SDFA}, we defined data movement as any data that must be sent outside of the neuron.
In the backwards pass this is the error information sent to each neuron in SSDFA and DFA. 
For BP implemented on a von Neumann architecture, this is equivalent to all the data that must be transported to and from main memory.
In Figures \ref{fig:ops}.B and \ref{fig:ops}.D we show the data movement results across SSDFA, DFA, and BP for CIFAR100 and ImageNet. 
% On the ImageNet dataset we observe a 1000$\times$ reduction in data movement from DFA to SSDFA and a 6400$\times$ reduction in data movement from BP to SSDFA.
On the ImageNet dataset we observe a 6400$\times$ reduction in data movement from BP to SSDFA.
This number reflects our approximations in Table \ref{tab:compare}, where BP requires $|W| + |A| + |E_{i+1}|$ words of data and SSDFA requires just a single error, $e$. 
From DFA to SSDFA, we observe a 1000$\times$ reduction in data movement since each neuron receives only a single error rather than 1000. 
% Given that each error requires only a single error signal, the communication becomes far more manageable compared with DFA and BP.
We show the data movement requirements for the CIFAR100 benchmark in Figures \ref{fig:ops}.C and \ref{fig:ops}.D.
We observe that the advantages of using SSDFA are less for smaller benchmarks because the number of errors and size of the hidden layers are less. 

\indent In Figure \ref{fig:ops}.A and \ref{fig:ops}.C we show the number of MACs across SSDFA, DFA, and BP for CIFAR100 and ImageNet. 
The reduction in the number of MACs from BP to SSDFA is nearly a factor of two. 
This reflects our approximations in Table \ref{tab:compare}, where we show that the number of MACs reduces from $2|W|$ to $|W| + |b|$. 
The number of MACs required to compute the error at a hidden layer reduces from $|W|$ to $|b|$, but the number of MACs to compute the partial error at each weight remains $|W|$.
Since we must compute the error at each feedforward weight, the reduction in MACs will always be bounded. 
If the feedforward matrix, $W$, was also sparse, then this bottleneck could be reduced and allow for a larger reduction in MAC operations. 

%%%%%%%%%%%%%%%%%%%%%%%%%%%%%%%%%%%%%%%%%%%%%%%%%%%%%%%%%%%%%%

\section{Discussion} \label {discussion}

\noindent 
Feedback Alignment and Direct Feedback Alignment have been proposed to address the weight transport problem in backpropagation. We propose Sparse DFA where the fixed feedback matrix used is constrained to be sparse. Such a sparse feedback matrix helps in a simpler physical implementation of communicating errors. To justify our arguments, we studied the training performance of networks MNIST, CIFAR10, CIFAR100, and ImageNet using feedback matrices with different constraints, such as rank and sparsity. We observe that rank of the feedback matrix has much stronger impact on accuracy, and making the feedback connections sparse has negligible effect on performance. Furthermore, using an extremely sparse version of SDFA where only a single error is fed back for weight update, we observe comparable performance while
minimizing data movement.

As was claimed in \cite{bartunov2018assessing}, Feedback Alignment, Direct Feedback Alignment, and as a result, the proposed Sparse Direct Feedback Alignment, do not scale to large networks. While our results are similar, we show that this is specifically true for convolutional networks due to the additional architectural constraints of repeated filters which they incorporate.
We show that by fixing the convolutional filters of the network to those trained using BP, and learning only the fully connected layers using DFA, we observe that the network's performance is close to that of BP. 
Our approach greatly simplifies the task of propagating the errors from the deeper end of the network to the shallow layers for learning the model weights. 

%%%%%%%%%%%%%%%%%%%%%%%%%%%%%%%%%%%%%%%%%%%%%%%%%%%%%%%%%%%%%%

\section{Methods} \label {methods}

\noindent We construct our networks in TensorFlow and create a new layer type so that we can feedback error directly to each hidden layer. 
To get optimal results we perform a hyper parameter search and also test different gradient descent optimizers and activation functions. We sweep learning rate and learning rate decay to find the optimal set. 
For weight initialization we used a uniform distribution in the range $ [-1 / \sqrt{fanout}, 1 / \sqrt{fanout}]. $
For feedback matrix initialization we used $ [-1 / \sqrt{fanin}, 1 / \sqrt{fanin}] $ where the input dimension was layer size. 
For sparse matrices we found $ [-1 / \sqrt{\frac{fanin \cdot N}{fanout}}, 1 / \sqrt{\frac{fanin \cdot N }{fanout}}] $ (where N is the number of sparse connections) to work well.

%%%%%%%%%%%%%%%%%%%%%%%%%%%%%%%%%%%%%%%%%%%%%%%%%%%%%%%%%%%%%%

\section*{Acknowledgements}
This work was funded by the U.S. Department of Defense’s Multidisciplinary University Research Initiatives (MURI) Program under grant number FOA: N00014-16-R-FO05 and the Semiconductor Research Corporation under the Center for Brain Inspired Computing (C-BRIC).

%%%%%%%%%%%%%%%%%%%%%%%%%%%%%%%%%%%%%%%%%%%%%%%%%%%%%%%%%%%%%%

\bibliographystyle{ieeetr}
\bibliography{main}

\end{document}